\documentclass{article}

\usepackage{PRIMEarxiv}

\usepackage[utf8]{inputenc} % allow utf-8 input
\usepackage[T1]{fontenc}    % use 8-bit T1 fonts
\usepackage{hyperref}       % hyperlinks
\usepackage{url}            % simple URL typesetting
\usepackage{booktabs}       % professional-quality tables
\usepackage{amsfonts}       % blackboard math symbols
\usepackage{nicefrac}       % compact symbols for 1/2, etc.
\usepackage{microtype}      % microtypography
\usepackage{lipsum}
\usepackage{fancyhdr}       % header
\usepackage{graphicx}       % graphics
\graphicspath{{media/}}     % organize your images and other figures under media/ folder

\usepackage{bbm}
\usepackage{multirow}
\usepackage{amsmath}

\title{The smooth output assumption, and why deep networks are better than wide ones
\thanks{\textit{\underline{Citation}}: 
\textbf{Authors. Title. Pages.... DOI:000000/11111.}} 
}

\author{
  Luis Sa-Couto, Jose Miguel Ramos, Andreas Wichert \\
  Department of Computer Science and Engineering and INESC-ID \\
  Higher Technical Institute, University of Lisbon \\
  Lisbon\\
  \texttt{\{luis.sa.couto, jose.miguel.ramos, andreas.wichert\}@tecnico.ulisboa.pt} \\
}

\begin{document}
\maketitle

\begin{abstract}
When several models have similar training scores, classical model selection heuristics follow Occam's razor and advise choosing the ones with least capacity. Yet, modern practice with large neural networks has often led to situations where two networks with exactly the same number of parameters score similar on the training set, but the deeper one generalizes better to unseen examples. With this in mind, it is well accepted that deep networks are superior to shallow wide ones. However, theoretically there is no difference between the two. In fact, they are both universal approximators.

In this work we propose a new unsupervised measure that predicts how well a model will generalize. We call it the output sharpness, and it is based on the fact that, in reality, boundaries between concepts are generally unsharp. We test this new measure on several neural network settings, and architectures, and show how generally strong the correlation is between our metric, and test set performance.

Having established this measure, we give a mathematical probabilistic argument that predicts network depth to be correlated with our proposed measure. After verifying this in real data, we are able to formulate the key argument of the work: output sharpness hampers generalization; deep networks have an in built bias against it; therefore, deep networks beat wide ones. 

All in all the work not only provides a helpful predictor of overfitting that can be used in practice for model selection (or even regularization), but also provides a much needed theoretical grounding for the success of modern deep neural networks.
\end{abstract}

% keywords can be removed

%%%%%%%
\section{Introduction}
During training, all a learning model can see is its training set. Yet, what it aims for is to generalize to unseen data \cite{Bishop:06}. 

When several models achieve similar training performances, the classical approach is to follow Occam's razor and use some kind of capacity penalizing heuristic to choose between them \cite{wichert:2021}.

Yet, modern Deep Learning practice has shown success with models that have an extremely high capacity but are constrained via regularization \cite{Belkin:19,sacouto:2022}.

For neural networks, the three most known kinds are: early stopping; random connection dropout during training \cite{srivastava2014dropout,Goodfellow:16}; and explicitly penalizing the $l_2$ norm of the weights \cite{Bishop:06}.

The latter is the only one that yields a metric that can stand as a predictor of overfitting. Its use can be justified either by the biological inspiration of weight decay \cite{Krogh:91}, or as an import from the math around nonlinear regression, where smaller $l_2$ norms favor smooth polynomials \cite{Bishop:06}. Having said that, as we will show in this work, it is merely an import, and thus not a very good predictor of overfitting for neural networks.

When two networks have the same number of parameters (i.e. capacity), and the same amount of regularization, deeper networks perform better \cite{Goodfellow:16,Belkin:19}. From a theoretical standpoint, this fact is very surprising given that one hidden layer is enough to make a network into a universal approximator \cite{cybenko1989approximation}. This advantage of more depth is so widespread that it is ultimately responsible for the popular name Deep Learning. And althought the field generally attributes it to gains of compositionality \cite{Goodfellow:16}, a theoretical grounding for it is still lacking.

Keeping this in mind, in section \ref{sec:hypo} we propose a new unsupervised predictor of how overfitted a neural network is, which is based on a heuristic knowledge of the boundaries between concepts in reality. After that, in section \ref{sec:experiments_corr} we test out our proposed measure on real data with lots of different architectures. Then, we analyze the consequences of such measure, and use section \ref{sec:experiments_depth} to build a theoretical and experimental argument that justifies why deep networks are better than wide ones.

\section{Output sharpness}
\label{sec:hypo}
When doing some machine learning task one can generally assume that the desired outputs will depend on a combination of input features. Typically, these features are numeric values, and small numeric changes to the input should not produce a large impact in the output. To this last sentence we call the unsharp output assumption. In a sense, it is based on how concepts in the world change, where we rarely have boolean logical definitions where am infinitesimal numerical change to a single feature should overturn the output completely.

Besides this intuitive origin, there is already machine learning precedence for this assumption. In a multivariate regression, $l_2$ regularization is typically used to ensure a smooth regression surface. Concretely, when the output is of the form of equation~\ref{eq:linear}, it is desirable that any given coefficient $w_i$ is small in absolute value. Otherwise, a small change $\Delta x_i$ to its feature, would have tremendous impact on the output value $o$.

\begin{equation}
    o=f\left(\mathbf{x}=\left[x_1,x_2,\cdots,_D\right]\right) = w_0 + x_1 w_1 + x_2 w_2 + \cdots + x_D w_D
    \label{eq:linear}
\end{equation}

In fact, across the most fundamental literature, when training a neural network with $l_2$ regularization, practitioners intend to achieve the so called output smoothness or unsharpness \cite{Bishop:06,Goodfellow:16}. However, in this work, we argue that this is not the best way to do so.

What we want is to see how much a small, possibly noisy, change to an input feature can overturn the output. So, we propose to compute the derivative of the output with respect to the input $\frac{\partial o}{\partial x}$ as an indicator of how smooth the model's decision. More concretely, we propose to use the norm of this derivative. A high value is a strong indicator of overfitting.

Let us take this idea and apply it to the simple multivariate regression case where the output can be computed with equation \ref{eq:linear}. In this case, the sharpness (or non-smoothness) derivative with respect to a given feature $x_i$ will be given by
\begin{equation}
\frac{\partial o}{\partial x_i} = w_i.    
\end{equation}
So, the gradient vector for all features will be
\begin{equation}
\nabla_{\mathbf{x}} o = \mathbf{w}.
\end{equation}
And so, the sharpness value, which is given by the norm of this derivative will yield
\begin{equation}
sharpness = \| \nabla_{\mathbf{x}} o \|_2 = \| \mathbf{w} \|_2 .
\end{equation}
Therefore, for the multivariate regression case, our proposed predictor recovers $l_2$ regularization theory. Now, for neural networks with hidden layers, instead of directly importing the usage of the norm, we argue that we should compute the model's sharpness as well.

Consider a general neural network with $L+1$ layers (including a fixed input layer $0$), where each layer $l$ contains $N\left(l\right)$ neurons. Assuming the input layer neurons are set to the input data values $a_i^{\left[ 0 \right]} = x_i$, the forward propagation equations can be written as follows, for $i=1,\cdots,N\left(l\right)$:
\begin{equation}
    z_{i}^{\left[ l \right]} = \sum_{k=1}^{N(l-1)} W_{ki}^{\left[ l \right]} a_{k}^{\left[ l-1 \right]}  + b_{i}^{\left[ l \right]},
\end{equation}
\begin{equation}
    a_{i}^{\left[ l \right]} = f^{\left[ l \right]} \left( z_{i}^{\left[ l \right]} \right).
\end{equation}
With this notation in mind, we can compute the sharpness derivative using the following recursion:
\begin{equation}
\beta_{ij}^{\left[ l \right]} = \frac{\partial a_{j}^{\left[ L \right]} }{ \partial a_{i}^{\left[ l \right]}  } = \left\{ \begin{array}{ll}
            \mathbbm{1}{\left[ i=j \right]} & l = L \\
             \sum_{k=1}^{N(l+1)} W_{ik}^{\left[ l+1 \right]} \frac{\partial a_{k}^{\left[ l+1 \right]} }{ \partial z_{k}^{\left[ l+1 \right]}  } \beta_{kj}^{\left[ l+1 \right]} & l < L
        \end{array} \right.
    \end{equation}
Now, this recursion is naturally given when we compute gradient backpropagation. To see this, note that, assuming a loss function $e = g\left( \mathbf{o}, \mathbf{y} \right) = g\left( \mathbf{a}^{\left[ L \right]}, \mathbf{y} \right)$, the gradient for a general weight can be written using the betas:
\begin{equation}
\frac{\partial e}{\partial W_{ij}^{\left[ l \right]} } = 
\frac{\partial z_{j}^{\left[ l \right]}  }{\partial W_{ij}^{\left[ l \right]}  } 
\frac{\partial a_{j}^{\left[ l \right]}  }{\partial z_{j}^{\left[ l \right]}  } 
\sum_{k=1}^{N(l)}
\frac{\partial a_{k}^{\left[ L \right]} }{ \partial a_{j}^{\left[ l \right]}  }
\frac{\partial e}{\partial a_{k}^{\left[ L \right]} } = 
a_{i}^{\left[ l - 1 \right]}
\frac{\partial a_{j}^{\left[ l \right]}  }{\partial z_{j}^{\left[ l \right]}  } 
\sum_{k=1}^{N(l)}
\beta_{jk}^{\left[ l \right]}
\frac{\partial e}{\partial a_{k}^{\left[ L \right]} }.
\end{equation}

So, in practice we require almost no added complexity to compute the desired derivatives. After backpropagating, we will have a beta for the first layer $\boldsymbol{\beta}^{\left[ 0 \right]} = \left[ \beta_{ij}^{\left[ 0 \right]}  \right]_{ij}$ which exactly the matrix of the derivatives of the outputs with respect to the inputs:
\begin{equation}
    \beta_{ij}^{\left[ 0 \right]} = \frac{\partial a_{j}^{\left[ L \right]} }{ \partial a_{i}^{\left[ 0 \right]} } = \frac{\partial o_j }{ \partial x_i }.
\end{equation}
Having this matrix, we can compute the sharpness of the network by computing its Frobenius norm
\begin{equation}
    sharpness = \| \boldsymbol{\beta}^{\left[ 0 \right]} \|_F
\end{equation}

\section{Sharpness predicts overfitting}
\label{sec:experiments_corr}
With our proposed metric clearly defined, it is time for us to test whether or not it predicts how overfitted a network is. 

Before moving to experimental tests we need to define the ideal conditions to test out our hypothesis. What we want is to be able to train many models in a realistic time frame, and to measure how well they generalize to unseen data. So, ideally, we want to have a training set that is relatively quick to learn, and a very large and representative test set.

To meet the first criterion, we took the $60000$ examples of the MNIST set of handwritten digits \cite{Lecun:98}, and downsampled the $28 \times 28$ images to a size $7 \times 7$. This new $49$ dimensional data set is much easier to learn, and thus, we can train many more networks.

To meet the second criterion, we can exploit the fact that we have an easier task and use only $1000$ training examples, thus saving the remaining $59000$ for testing. All our networks were trained for a large number of epochs ($5000$) using stochastic gradient descent with typical momentum values, and a decaying learning rate.

Having gone through the technical details, let us start by looking at a single linear layer, which is equivalent to a multivariate regression.

\subsection{Linear models and the weight norm}
\begin{figure}
\centering
\includegraphics[width=0.45\textwidth]{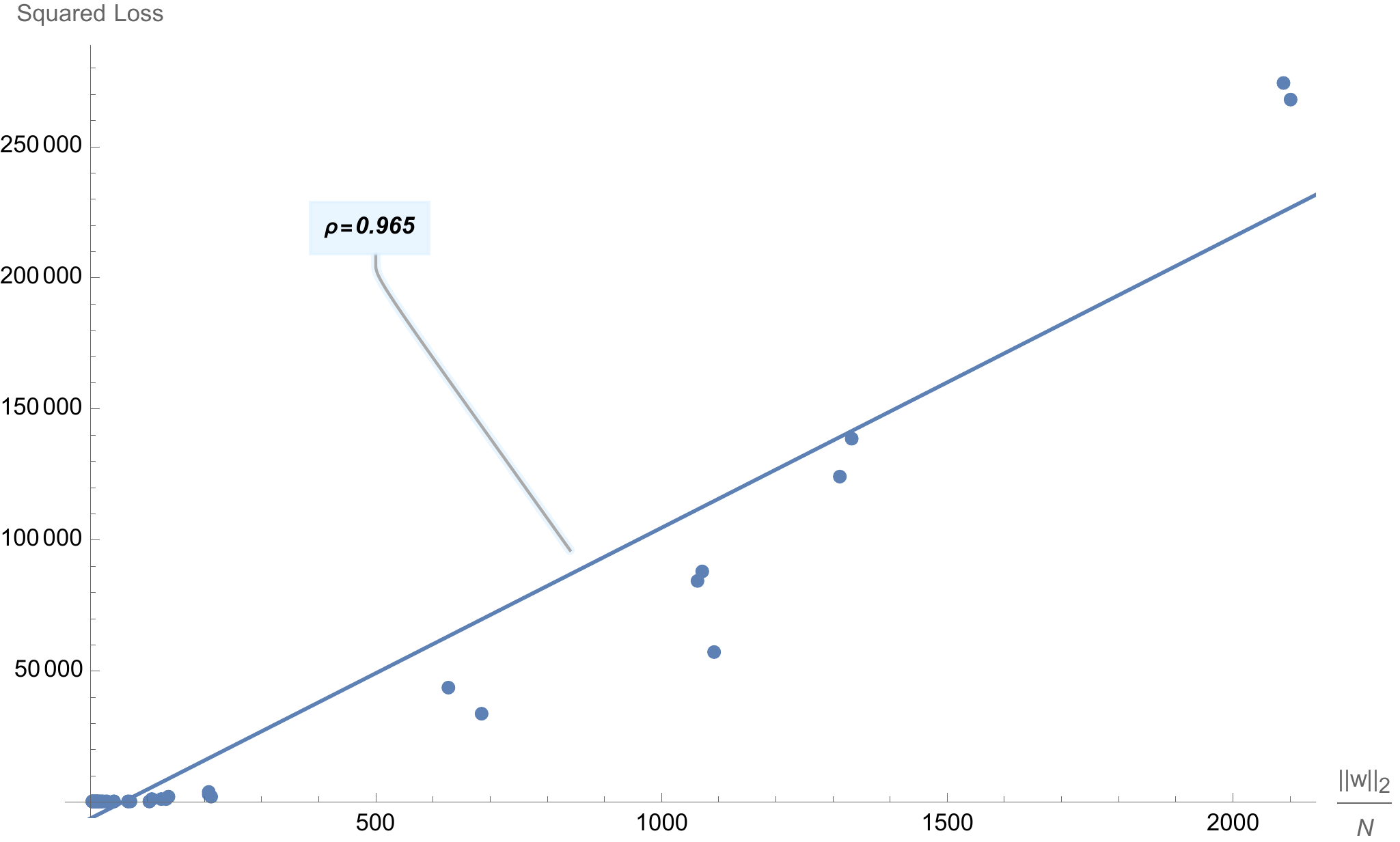}
\caption{The horizontal axis contains the normalized norm of the weights which is equivalent to the model output sharpness. The vertical axis contains the squared loss values measured on the big test set. Each point is a trained network, and we can see that, as expected, there is a very strong correlation between sharpness and test loss.}
\label{f1}
\end{figure}

As we have shown in the previous section, for the linear model sharpness equals the $l_2$ norm of the weights. This norm is typically used in regularization (alongside $l_1$ norm for sparse solutions \cite{Bishop:06}), and so we expected it to be a great predictor of squared loss. To verify this we took four random ReLU feature transformations with dimensions $300$, $800$, $1200$ and $1800$, and apply them to the training set. Then, for each feature transformation, we learned the optimal weights using the closed form solution for squared error. Given that all listed transforms make the problem over-parameterized (with $300$ features, the model already has $3000$ parameters), the learning problem has many equivalent solutions. For each run, we choose the solution that is closest to a given random starting point with one of many possible norms between $0.1$ and $1000$. In total, we trained $57$ linear networks on the $1000$ training examples. Having the trained models we measured their sharpness, which in this case is the norm of the weights, and computed their performance on the big test set. These results are plotted in figure \ref{f1} where each point represents a network. Analyzing the results, we see that there is a very strong correlation of $0.965$ between the normalized norm of the weights and the squared error.

\begin{figure}
\centering
\includegraphics[width=0.45\textwidth]{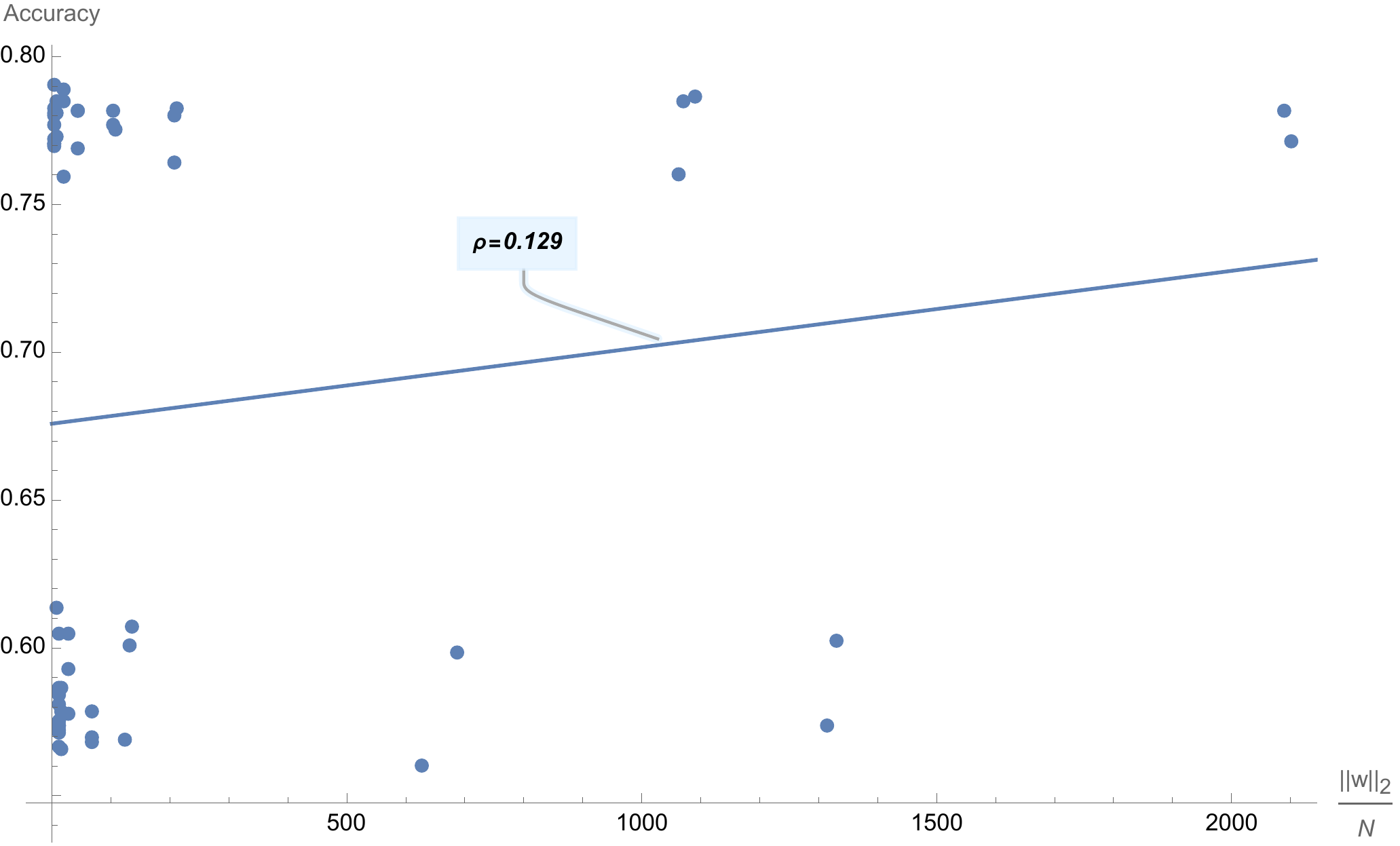}
\caption{The horizontal axis contains the normalized norm of the weights, and the vertical one contains test accuracies. Each point is a trained network, and we can see that there is not a clear relationship between $l_2$ norm and test accuracy.}
\label{f2}
\end{figure}

\begin{figure}
\centering
\includegraphics[width=0.45\textwidth]{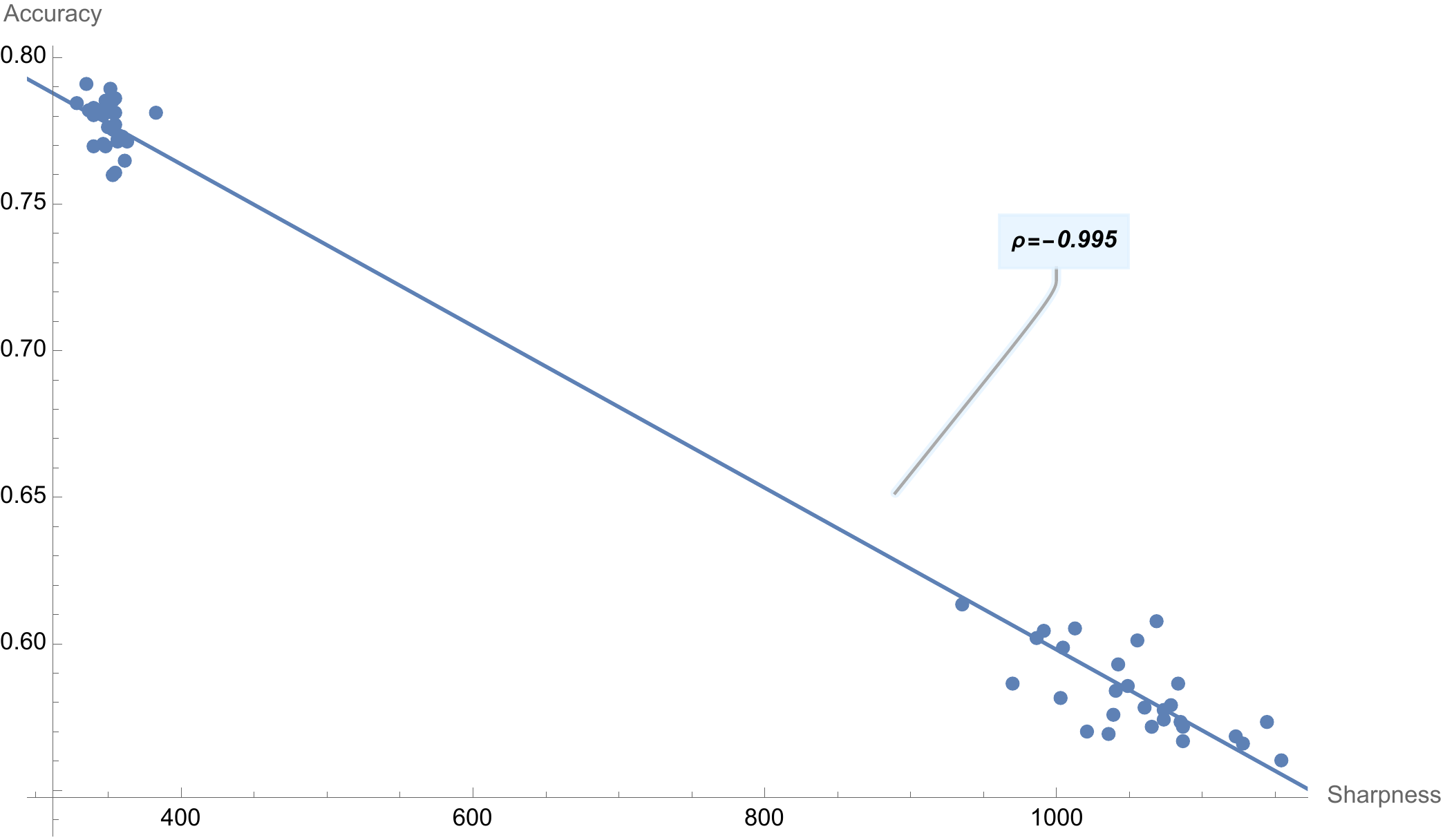}
\caption{The horizontal axis contains the sharpness assuming a softmax activation function, whereas the vertical axis contains test accuracies. Each point is a trained network, and we can see an incredibly strong correlation between the two variables.}
\label{f3}
\end{figure}

If we take all the networks, and compute their test accuracies instead, we get the results plotted in figure \ref{f2}. In this case, the norm is not a good predictor of overfitting. But what about the sharpness? Is it just as bad? To compute it in this case we must look at the model differently. Concretely, it is now outputting class labels instead of regression numbers. We need to consider this and change our derivative. A valid example would be to look at all the networks in our experiment as if they had a softmax activation function. By computing the sharpness of that version, we get the results presented in figure \ref{f3}. We get an incredible predictor of overfitting in terms of test set accuracy: the correlation between sharpness and test set accuracy is $-0.995$.

These preliminary results seem to confirm our intuition that overfitting on a given task is very related to output sharpness as we have defined it. But do these intuitions hold for networks with many layers?

\subsection{Sharpness with full fledged networks}
\begin{figure}
\centering
\includegraphics[width=0.45\textwidth]{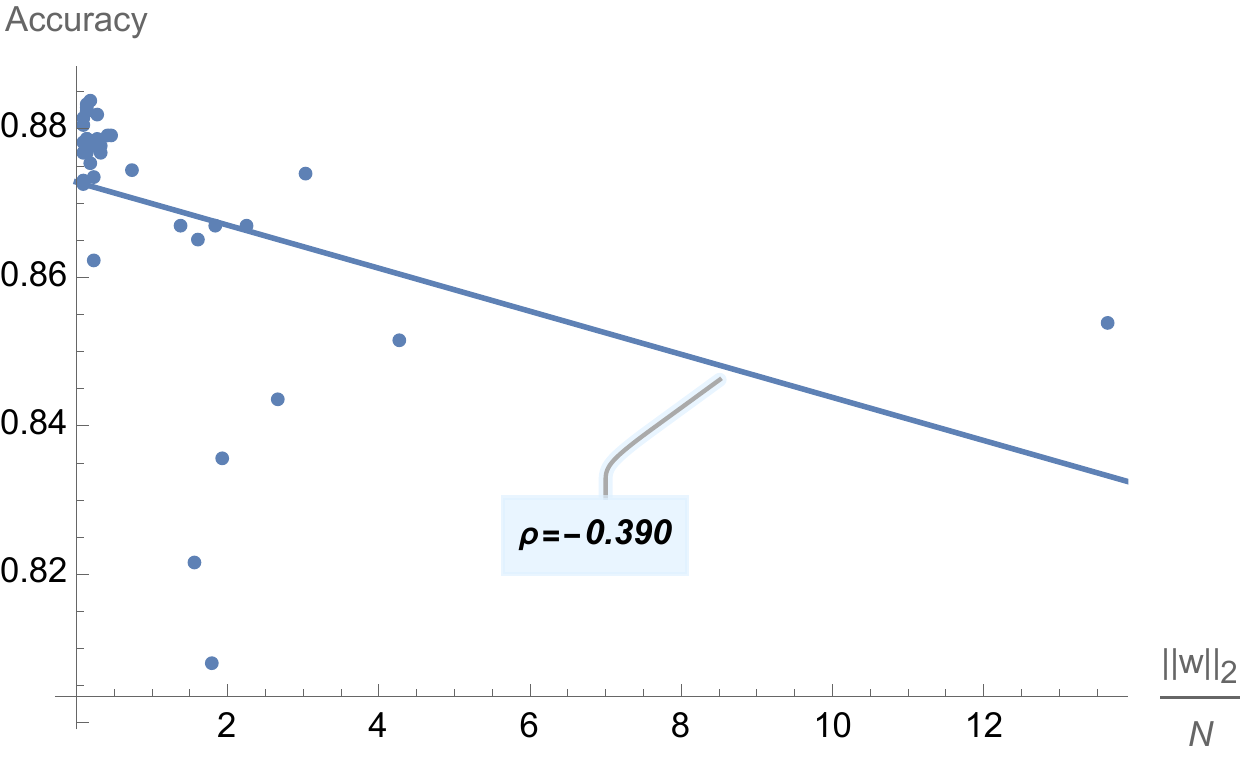}
\includegraphics[width=0.45\textwidth]{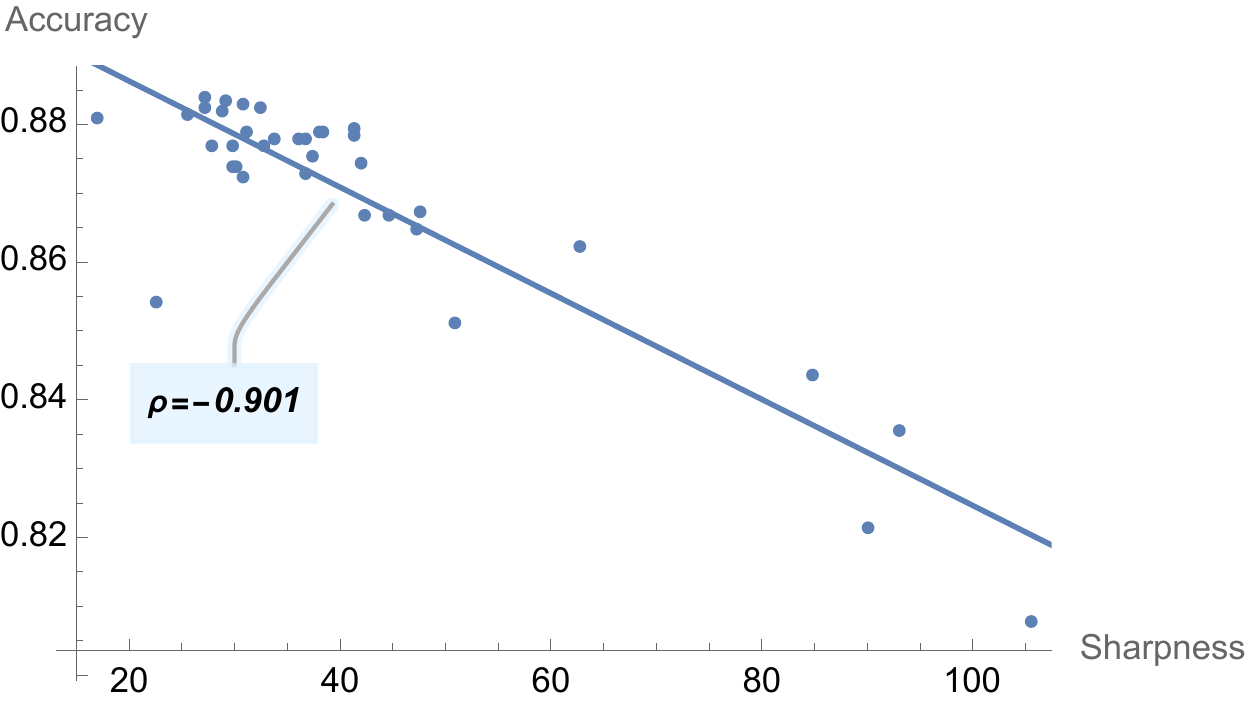}
\caption{For each of the $36$ trained networks we computed the weight norm, the sharpness, and the test accuracy. Using these three quantities we build two scatter plots. On the left we see norm vs. accuracy, whereas on the right we see sharpness vs. accuracy. Though both seem to be interesting predictors of overfitting, our proposed measure shows a much stronger relation.}
\label{f4}
\end{figure}

To start our search for validation, we constructed a large family of neural networks with tanh hidden units, and softmax output units (the chosen loss was categorical crossentropy). We construct networks with several different numbers of parameters between $1000$ and $14000$. Then, we uniformly distribute the resulting neurons across a number of hidden layers that can go between $1$ and $6$ (the approximate number of units per layer is presented in table \ref{tab:archi}). In total, we build and fully train $36$ networks. For each trained network, we compute three quantities: the average weight norm; the output sharpness; and the test set accuracy. The results are presented in the two scatter plots of figure \ref{f4}. Analyzing them we see that although the $l_2$ norm of the weights has a relation to test set accuracy (correlation of $-0.390$), our proposed measure has a tremendously clearer relation with a very high correlation of $-0.901$.

\begin{table}[h]
    \centering
\begin{tabular}{|c|c c c c c c||} 
\hline
  & 1000 & 5000 & 8000 & 10000 & 12000 & 14000\\ [0.5ex] 
\hline\hline
1 & 17 & 84 & 134 & 167 & 200 & 234 \\
\hline
2 & 14 & 47 & 64 & 75 & 84 & 92 \\
\hline
3 & 12 & 37 & 50 & 57 & 64 & 70 \\
\hline
4 & 11 & 32 & 42 & 48 & 54 & 59 \\ 
\hline
5 & 10 & 28 & 37 & 43 & 47 & 52\\
\hline
6 & 9 & 26 & 34 & 39 & 43 & 47\\
\hline
\end{tabular}
    \caption{For each desired approximate number of parameters, neurons are distributed uniformly across the network's hidden layers. Parameters can go from 1000 to 14000, whereas hidden layers range from 1 to 6. The table contains the approximate number of neurons per layer in each of the 36 trained networks. For example, entry (4,8000) contains the value 42. This means that we trained a network with 4 hidden layers where each one had around 42 units making a total of 8000 parameters.}
    \label{tab:archi}
\end{table}

\begin{figure}
\centering
\includegraphics[width=0.45\textwidth]{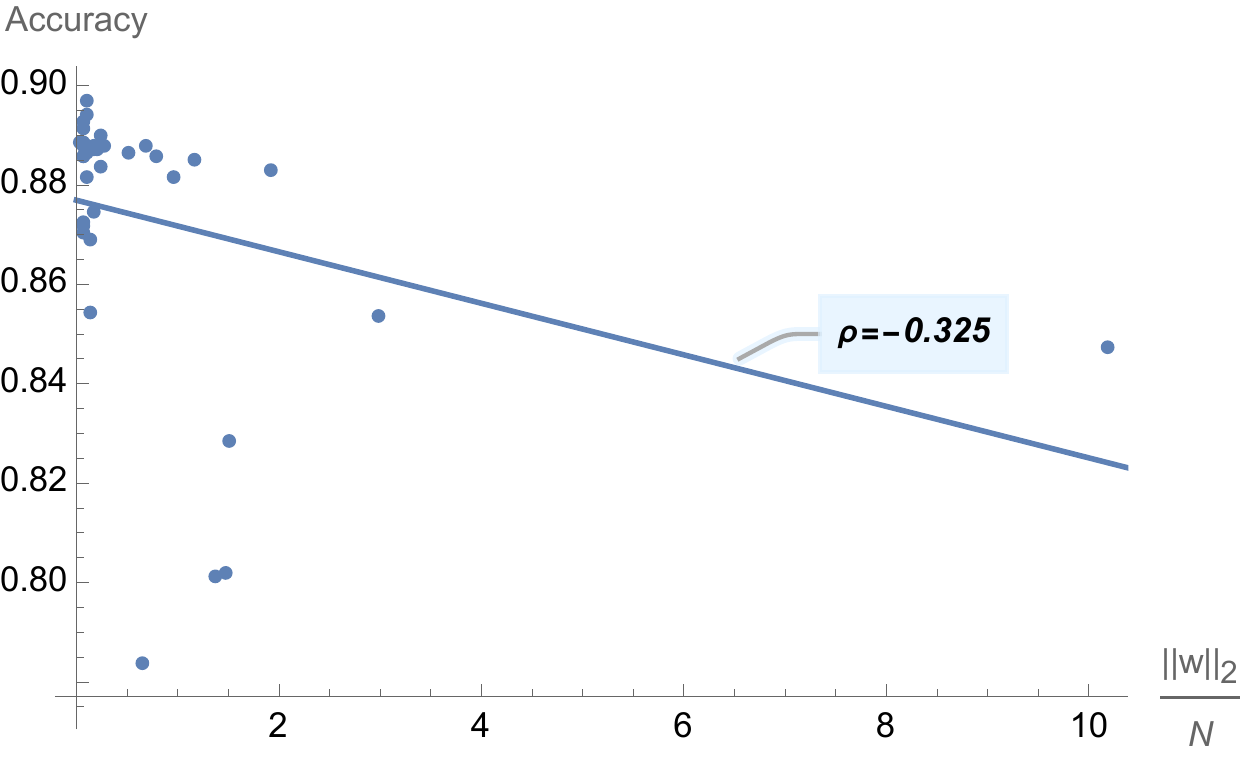}
\includegraphics[width=0.45\textwidth]{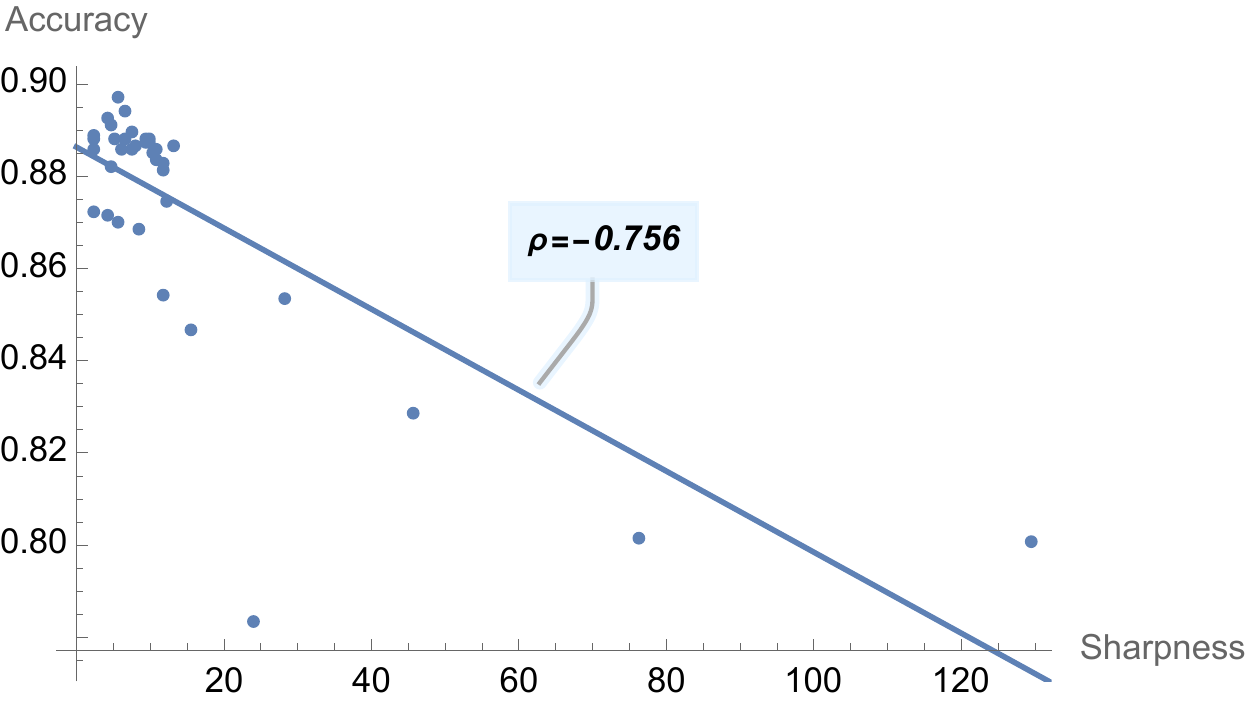}
\caption{For each of the $36$ trained ReLU-Softmax-Crossentropy networks we computed the weight norm, the sharpness, and the test accuracy. Using these three quantities we can build two scatter plots. On the left we have norm vs. accuracy, whereas on the right we have sharpness vs. accuracy. Both relate to overfitting. However, sharpness exhibits a much stronger anticorrelation of $-0.756$.}
\label{f6}
\end{figure}

The results are once again very encouraging. However, we need to make sure that they are not unique to Tanh-Softmax-Crossentropy networks. To answer this question we repeated the exact same experimental protocol as before, but this time our hidden activation funtion was ReLU. The achieved results are presented in figure \ref{f6}, and, once again they seem to point in the same direction.

\begin{figure}
\centering
\includegraphics[width=0.45\textwidth]{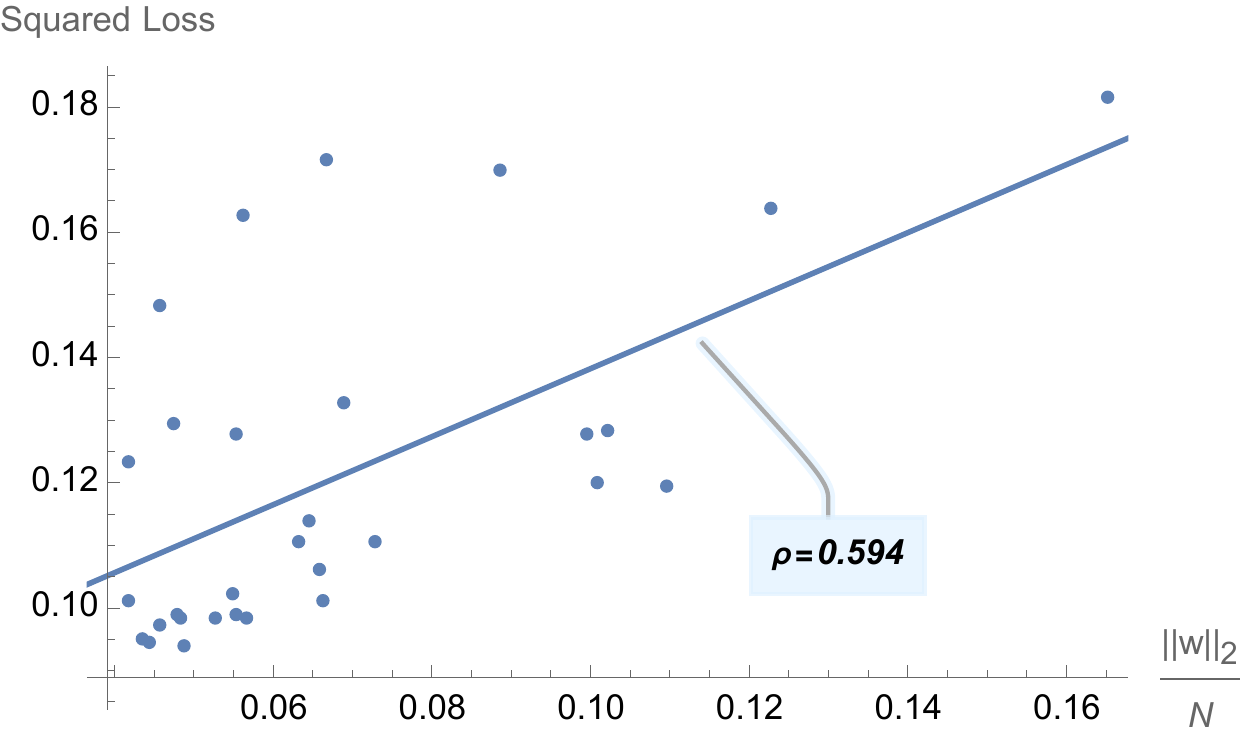}
\includegraphics[width=0.45\textwidth]{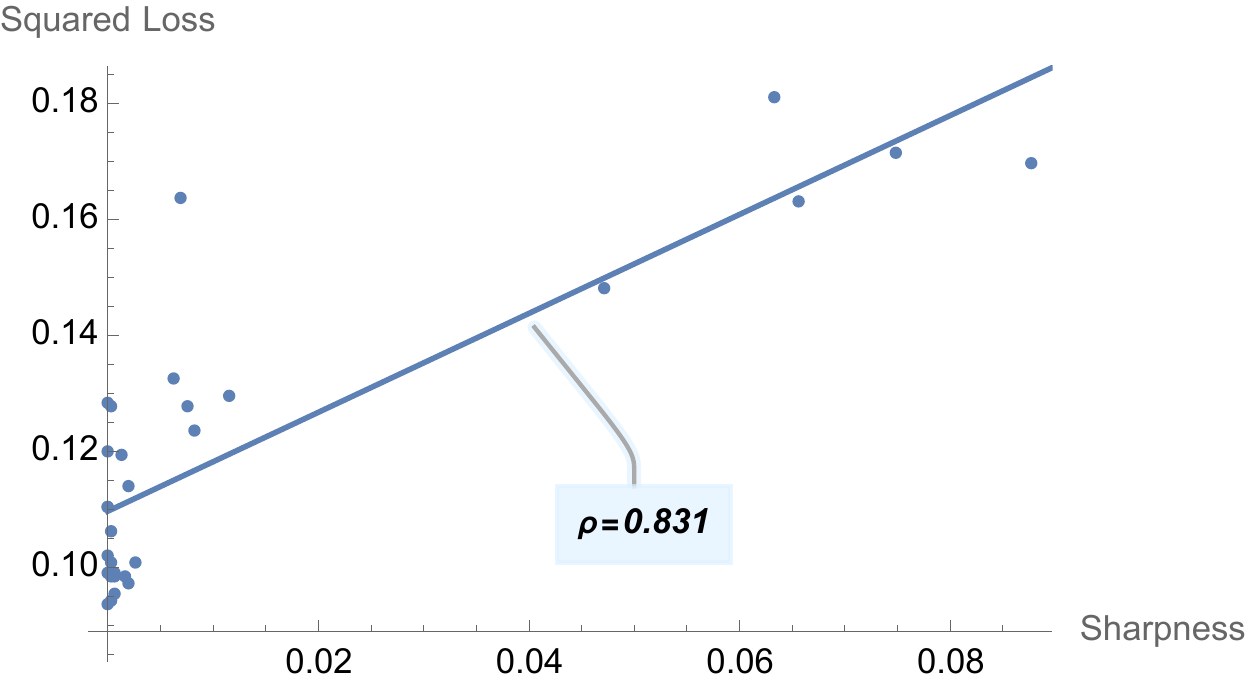}
\caption{For each trained ReLU-Linear-SquaredError network we computed the norm, the sharpness, and the test loss. Using these quantities we built two scatter plots. On the left we have norm vs. loss, whereas on the right we have sharpness vs. loss. Both relate to overfitting. However, sharpness exhibits a much stronger correlation of $0.831$.}
\label{f8}
\end{figure}

With comprehension as our goal, we repeated the same experiments but for a squared error loss function. This time, output units have linear activation functions, and we control the squared loss instead of accuracy. The results are depicted in figure \ref{f8}, and although the norm seems to be a better predictor than before, sharpness is still clearly better.

In summary, our proposed measure seems to truly capture overfitting with an almost linear relationship. That alone is enough to make it useful. Especially if one bears in mind that in our experiments the test set contains almost all the data, so our measure is predictive of more than a mere validation score. Hence, model selection based on sharpness can be done without any validation data, thus avoiding validation set overfitting, and also avoiding the sacrifice of valuable training data.

\section{Depth and sharpness}
\label{sec:experiments_depth}
Having established sharpness as a good indicator of overfitting, we noticed an interesting relation between it and a network's depth. More specifically, if we recall section \ref{sec:hypo}, and write the gradient for a weight on the first layer
\begin{equation}
\frac{ \partial e }{ \partial W_{ik}^{ \left[ 1 \right] } } = 
\sum_{k=1}^{N(1)}
\frac{\partial z_k^{ \left[ 1 \right]}}{\partial  W_{ik}^{ \left[ 1 \right] } }
\frac{\partial a_j^{ \left[ L \right]}}{\partial z_k^{ \left[ 1 \right]}}
\frac{\partial e}{\partial a_j^{ \left[ L \right]}}
=
x_i
\sum_{j=1}^{N(L)}
\frac{\partial e}{\partial a_j^{ \left[ L \right]}}
\sum_{k=1}^{N(1)}
\frac{\partial a_j^{ \left[ L \right]}}{\partial z_k^{ \left[ 1 \right]}},
\end{equation}
and also write the sharpness for a given input-output combination
\begin{equation}
\beta_{ij}^{ \left[ 0 \right] } = 
\sum_{k=1}^{N(1)}
\frac{\partial z_k^{ \left[ 1 \right]}}{\partial a_i^{ \left[ 0 \right]}}
\frac{\partial a_j^{ \left[ L \right]}}{\partial z_k^{ \left[ 1 \right]}}
=
\sum_{k=1}^{N(1)}
W_{ik}^{ \left[ 1 \right] }
\frac{\partial a_j^{ \left[ L \right]}}{\partial z_k^{ \left[ 1 \right]}},
\end{equation}
we see that there is a relation between output sharpness, and the gradients that reach the first layer through the terms $\frac{\partial a_j^{ \left[ L \right]}}{\partial z_k^{ \left[ 1 \right]}}$.

Being made of a chain of products of activation function derivatives, which are usually values between zero and one, these gradients get smaller as the network gets deeper. So much so, that the well known phenomenon of vanishing gradients can hamper learning if no care is taken \cite{Goodfellow:16}.

So, if network depth will reduce the absolute value of these gradients, it will also reduce the absolute value of the sharpness matrix values, thus reducing the matrix norm. This tells us that as depth grows sharpness will tend to decrease. Intuitively this makes sense since many transformations happen between the input and the output, so it is only natural that small changes at the start will have a reduced impact at the end.

\begin{figure}
\centering
\includegraphics[width=0.45\textwidth]{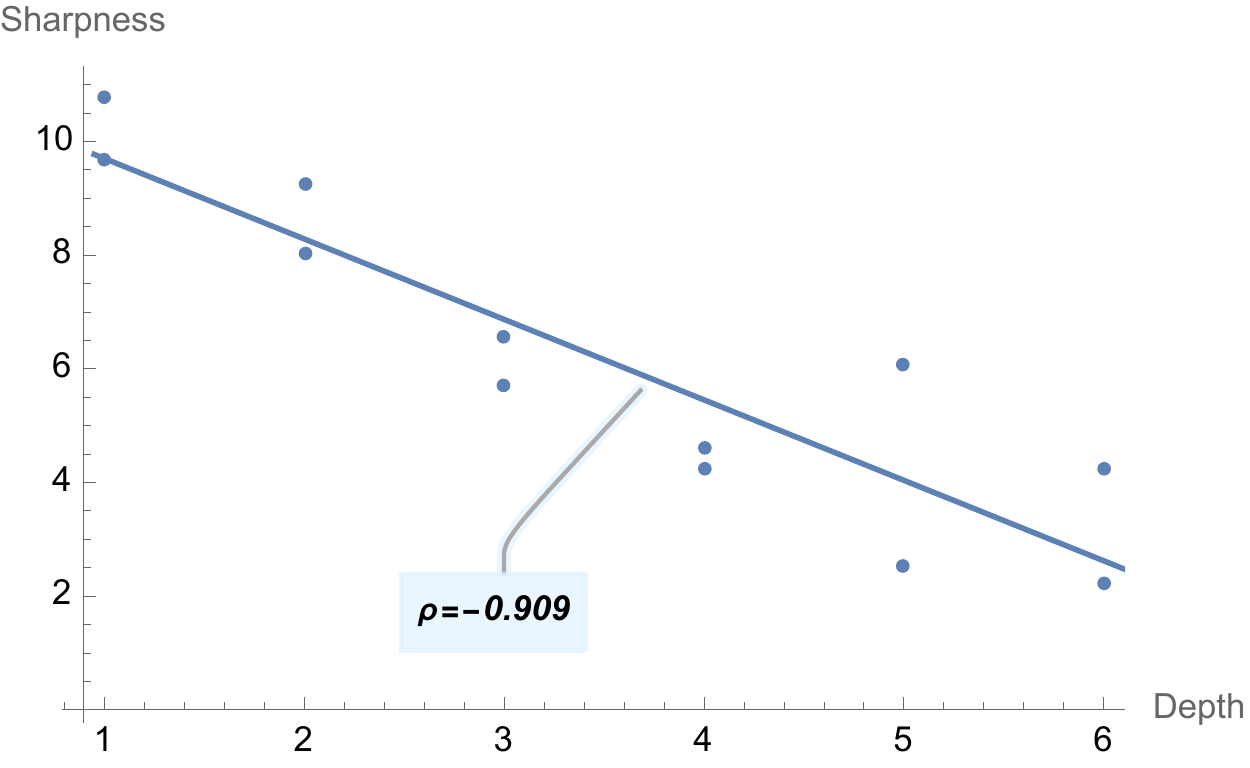}
\includegraphics[width=0.45\textwidth]{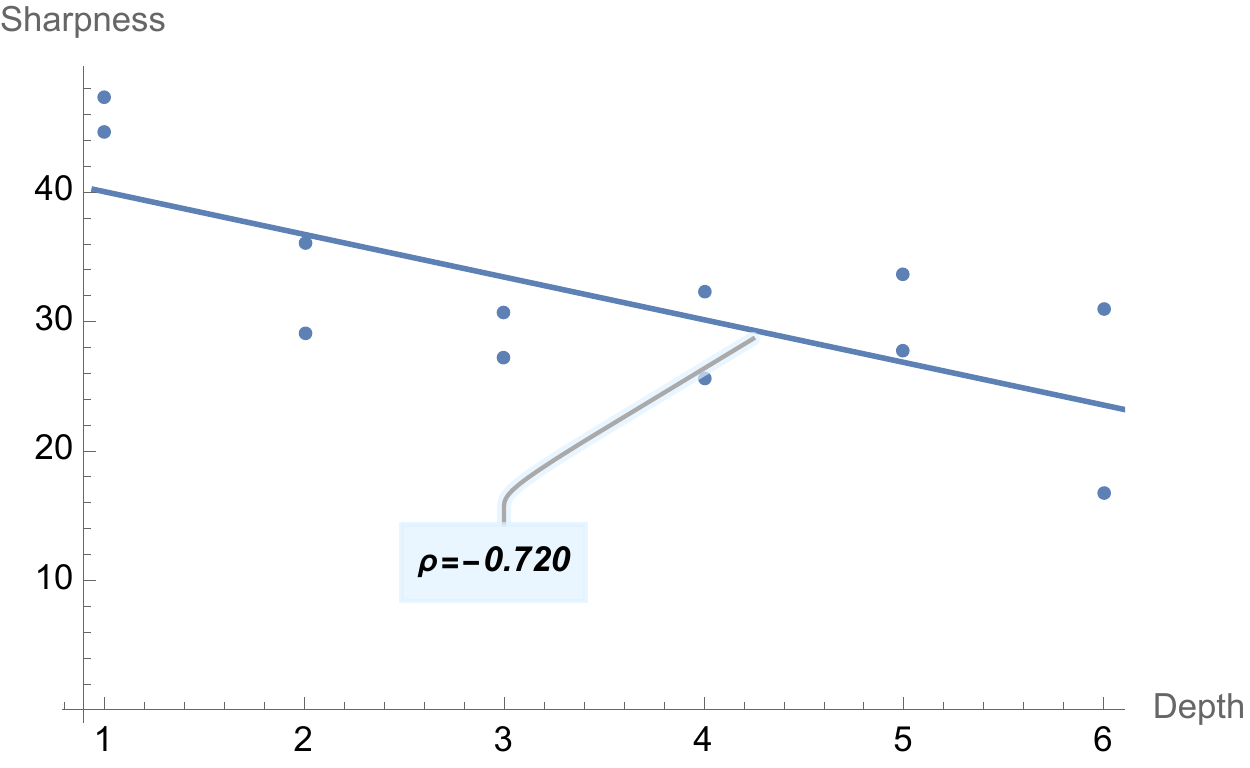}
\includegraphics[width=0.45\textwidth]{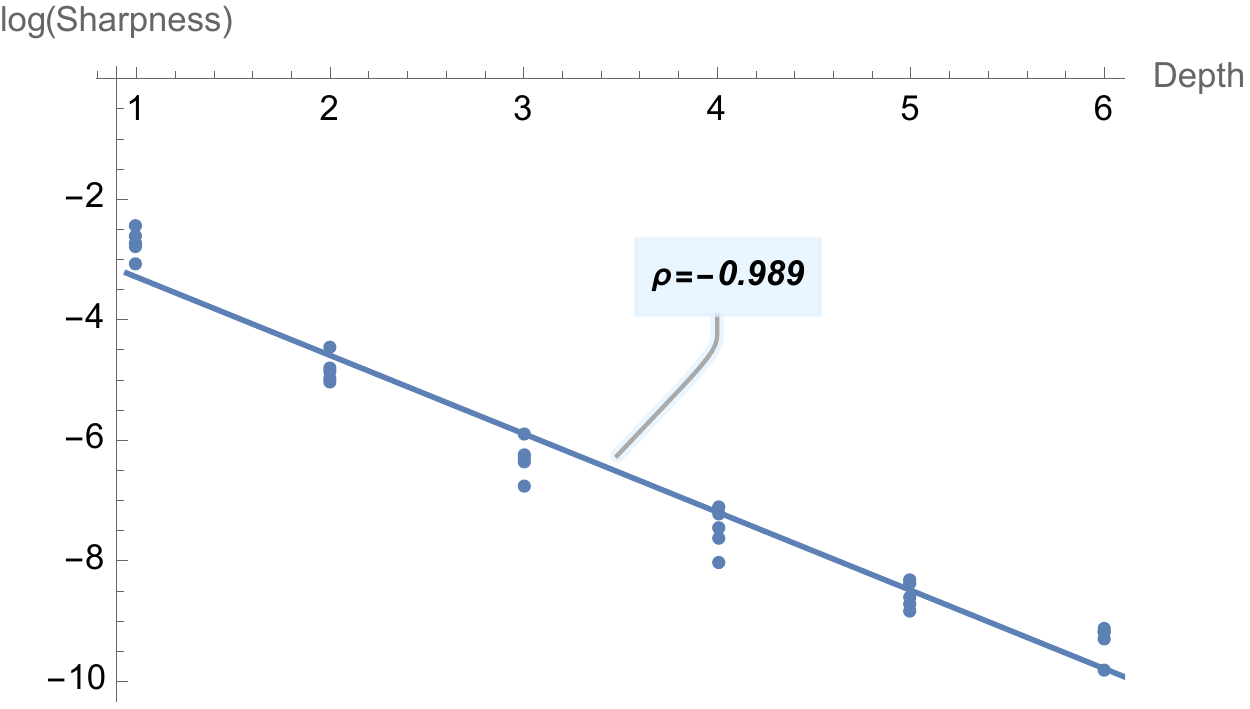}
\caption{Three scatter plots were made to illustrate the relation between sharpness and network depth. On the top we have the crossentropy architectures and the relation between the two variables seems to be strongly linear. On the bottom we present the squared error (with linear output neurons) networks where the relation between the variables is clearly logarithmic.}
\label{f10}
\end{figure}

To validate this prediction, we took all the networks from the previous section and created scatter plots of depth vs. sharpness. The results are depicted in figure \ref{f10}, and they stand in clear favor to our prediction.

\section{Conclusion}
When a model is not overfitted to training data, small noise in an input feature should not change the output by a lot. That is, the model's output should be smooth, or unsharp. Based on this assumption, we proposed a new unsupervised measure of overfitting for neural networks: the output sharpness.

This measure is conveniently computed with gradients that we already have to compute when doing backpropagation for stochastic optimization. And our experiments have shown that it predicts generalization error better then alternative criteria like the $l_2$ norm of the weights. We have shown this on many different architectures, both shallow and deep, using classification and regression. Furthermore, our generalization scores are measured on a very large data set, and not on a small validation set, so the results are significant.

Having established this measure, we gave a probabilistic argument that is based on a softer version of the well known phenomenon of vanishing gradients. Concretely, it predicted that as the network's depth increases, sharpness should decrease. Or, put differently that deep networks have an in built bias against output sharpness. We took the same broad set of trained networks and verified our predictions on real data.

In summary, we proposed a helpful predictor of overfitting that can be used in practice for model selection (or even regularization). Additionally, if we take it as a true measure of generalization, we can theoretically see why deep networks outperform wide ones. A theoretical insight that has been long sought after.

\section*{Acknowledgments}
We would like to acknowledge support for this project
from the Portuguese Foundation for Science and Technology (FCT) with a doctoral grant SFRH/BD/144560/2019 awarded to the first author, and the general grant UIDB/50021/2020. The Foundation had no role in study design, data collection and analysis, decision to publish, or preparation of the manuscript. The authors declare no conflicts of interest. Code and data for all the experiments can be obtained by email request to the first author.

%Bibliography
\bibliographystyle{unsrt}  
\bibliography{paper}

\end{document}